# Towards a Generic Framework for the Development of Unicode Based Digital Sindhi Dictionaries

IMDAD ALI ISMAILI*, ZEESHAN BHATTI** AND AZHAR ALI SHAH**

## ABSTRACT

Dictionaries are essence of any language providing vital linguistic recourse for the language learners, researchers and scholars. This paper focuses on the methodology and techniques used in developing software architecture for a UBSESD (Unicode Based Sindhi to English and English to Sindhi Dictionary). The proposed system provides an accurate solution for construction and representation of Unicode based Sindhi characters in a dictionary implementing Hash Structure algorithm and a custom java Object as its internal data structure saved in a file. The System provides facilities for Insertion, Deletion and Editing of new records of Sindhi. Through this framework any type of Sindhi to English and English to Sindhi Dictionary (belonging to different domains of knowledge, e.g. engineering, medicine, computer, biology etc.) could be developed easily with accurate representation of Unicode Characters in font independent manner.

Key Words:   Sindhi-to-English Dictionary, English-to-Sindhi Dictionary, Digital Dictionary.

## 1.    INTRODUCTION

ICT (Information and Communication Technologies) have changed almost all the traditional landscapes. Like all other domains, ICT has also become an integral and essential part of our educational system. Be it scientific management or online provision of services, ICT has changed the ways of research, teaching, learning and social networking. This paradigm shift is making the system more optimal, efficient and convenient. It is under this influence that each country is reshaping the provision of its public services (education, health, security etc.). This reshaping could be visualized, in the context of Pakistan, by taking example of its National Digital Library (http://www.digitallibrary.edu.pk/ ). This library provides online (electronic) delivery of about 75,000 academic resources in the form of eBooks, journals and conference articles, dissertations, reports, databases etc. In order to enrich such a library with the contents written in our national and sub-national languages, there is a dire need for the development of more and more digital resources. In this context, this paper introduces a novel generic framework for the development of Unicode based digital Sindhi dictionaries. Sindhi is the official language of Sindh and is regarded as one of the oldest spoken languages of the world [1-3]. Though, there have been some efforts for the computerization of standard Sindhi dictionary [4-5], but their approach is based on third party software and is fixed to only one type of dictionary.  This paper presents a generic software framework that is not based on any third-party software and could be used on any computer platform to easily develop any type of digital Sindhi dictionary. The development of such dictionaries and their provision through Pakistan's National Digital Library would highly facilitate the research work on Sindhi language, literature and linguistics. Needless to say that the dictionary or a lexicon provides definitions, usage, etymologies, phonetics, pronunciations and other information in one language with their equivalents in another [6]. Thus, a great source of learning, research and development of a language providing vital resources for the development of various computerized systems including spell checkers, machine translations, corpus building and analysis, text to speech and speech to text systems to name a few. The organization of the rest of this paper is as follows: Section 2 presents are overview of the proposed software architecture for this work; Section 3 describes the GUI (Graphical User Interface) and the workflow of the system; and finally, Section 4 concludes the article.

\*        Professor and Pro Vice-Chancellor, University of Sindh, Khairpur Mirs Campus.
\*\*      Assistant Professor, Institute of Information & Communication Technologies, University of Sindh, Jamshoro.

## 2. SOFTWARE ARCHITECTURE

The software architecture for our proposed system is based on Java technology. The use of Java technology provides platform independence and supports the Unicode character encoding scheme very efficiently. The framework employs Hash table structure of Java for the efficient storage and retrieval of the dictionary items as opposed to conventional approaches which make use of third-party databases. The framework also makes use of standard Java input/output operations for writing/reading data to/from a file as a serialized object. Further details of the framework are presented in the following subsections.

### 2.1 Hash Table and Hash Function

Hash table or a hash map is a data structure that stores the data in the form of 'key' and 'value' pairs. The key (e.g. any Sindhi word) is used as an identifier and is associated with a corresponding value (e.g. the English meaning of the Sindhi word). The association between the key and its value is performed by a mathematical function called 'hash function'. Based on the contents of the 'key', the hash function uses a mathematical expression to calculate a unique index which points to the element (slot or bucket) of the hash table that stores the corresponding 'value'. The main purpose of using hash tables over other data structures is speed. This advantage is more apparent when the number of entries is large (thousands or more) as aimed in our project to incorporate at least more than 50,000 words in the dictionary. The other reason is that the hash tables are particularly efficient when the maximum number of entries can be predicted in advance, so that the bucket array can be allocated once with the optimum size.

The hash table structure used in our dictionary is shown in Figs. 1-2. It employs Sindhi word as the key identifier for Sindhi to English dictionary and English word in the case of English to Sindhi dictionary. For its associated element we created a custom java object to save related information such as pronunciation, grammatical structure and meaning. The details of this object are presented in the following subsection.

### 2.2 Managing the Dictionary Data

Fig. 3 shows the user activity diagram in the context of updating the contents of the dictionary. Each dictionary entry consists of five fields i.e. word, pronunciation, grammar, Sindhi meaning and English meaning. Once user fills in these fields and directs the system to store the record, then the system puts this record into the hash table. Each hash table entry is in the form of 'key' and 'value' (element). We implemented a Java class to specify the contents of the 'value' as an object (DictinayData Object). Fig. 4 shows the class diagram of the DictionaryData class. This class contains the methods for storing and retrieving record for each word from the object which will be saved in a hash table with corresponding word as key. The methods starting with 'set' e.g. setPronc(), setGrammer(), setSindhiMeaning() etc have been implemented for constructing the dictionary, while the methods starting with 'get' e.g. getPronc(), getGrammer(), getSindhiMeaning() have been implemented to provide access to the entries of the dictionary. This same class would be used for both categories of dictionary that is Sindhi to English and English to Sindhi Dictionary.

*Towards a Generic Framework for the Development of Unicode Based Digital Sindhi Dictionaries*

### 2.3 Tokenizing Sindhi Meaning into Multiple Sindhi Words

One major methodology that has been formulated in this system is to provide a single input mechanism for both categories of dictionaries. That is, when the user opts for English to Sindhi dictionary as its primary input source, the Sindhi meaning entered for each English word is tokenized/separated into multiple Sindhi words and the information is saved in Sindhi to English dictionary with its relevant DictionaryData object. Through this approach user only needs to enter a single record for English to Sindhi dictionary, and the Sindhi to English dictionary is automatically created by the system. This procedure of tokenizing Sindhi meaning into multiple Sindhi words is illustrated in Fig. 5.

### 3. GUI and the Workflow of the System

The system provides the functionality for the user to opt which dictionarys/he wants to create or work with (Fig. 6). The main interface initially provides the list of words previously entered into the database or will be blank if no word has been saved (Fig. 7). When a word is selected from the list through mouse, then its corresponding data is displayed on the relevant fields which initially are not editable. The GUI provides the features for adding new records, editing previously saved records and deletion of any incorrect record. These features are described in the following subsections.

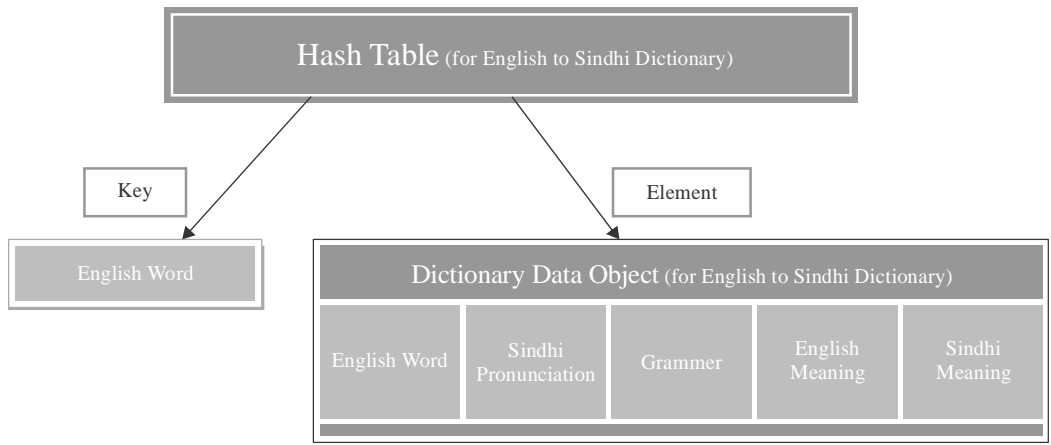

FIG. 1. HASH TABLE STRUCTURE FOR THE ENGLISH TO SINDHI DICTIONARY

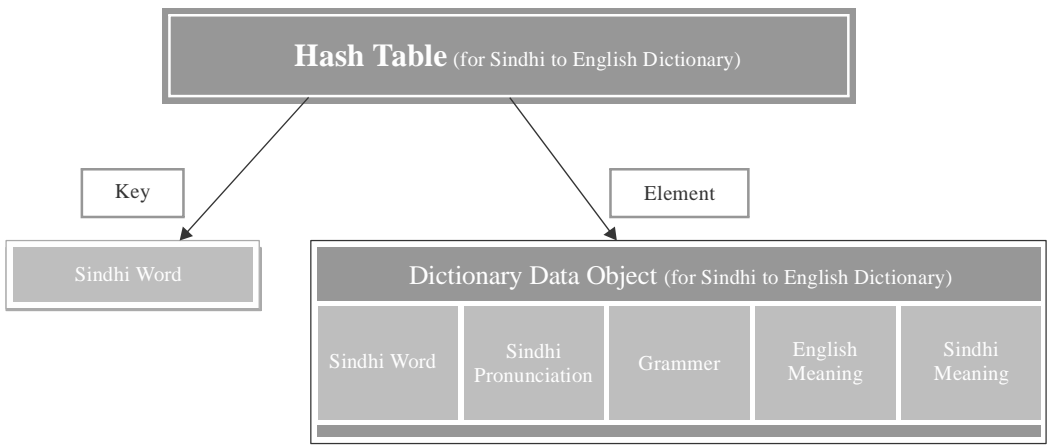

FIG. 2. HASH TABLE STRUCTURE FOR THE SINDHI TO ENGLISH DICTIONARY

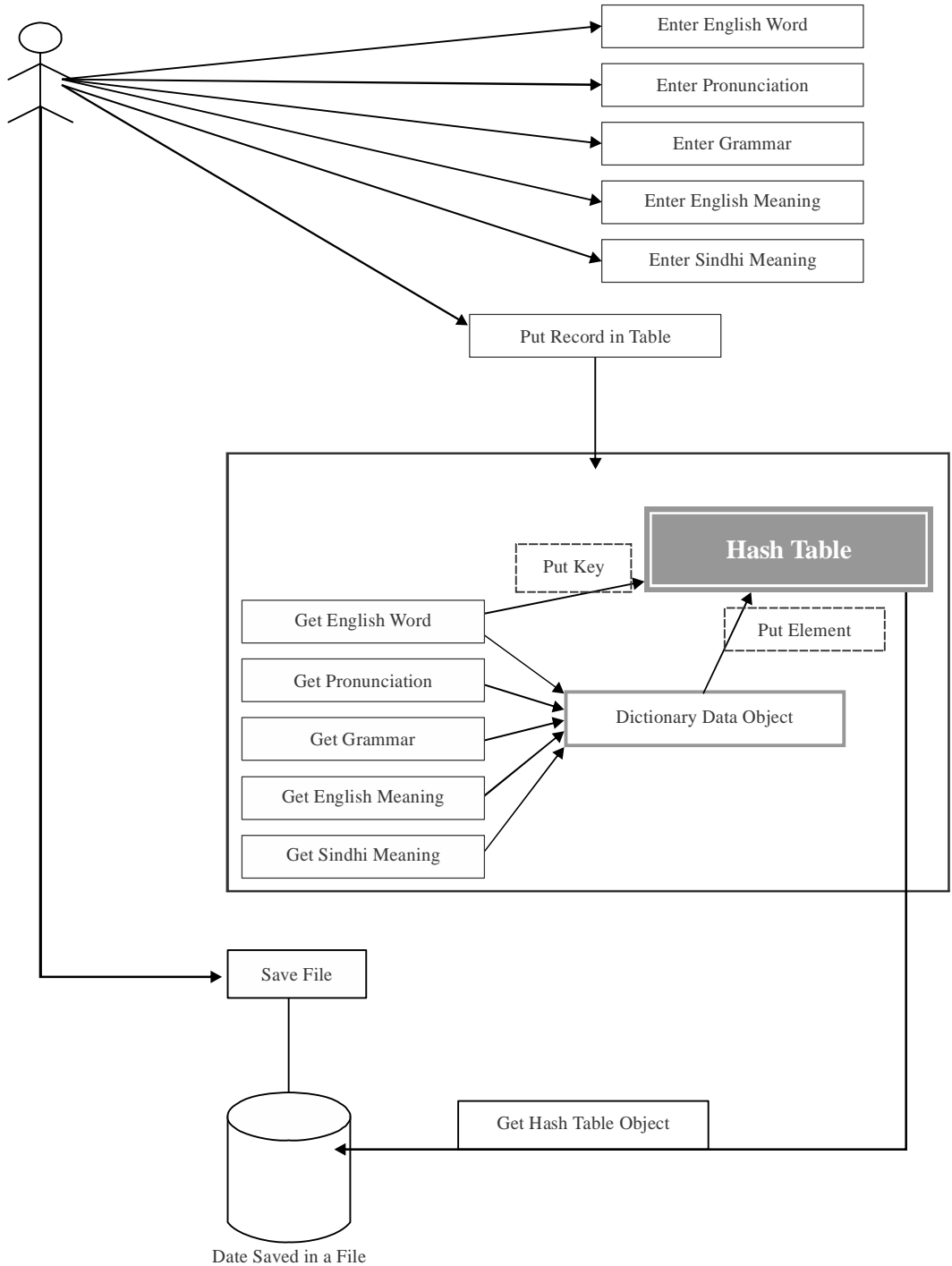

*FIG. 3. USER ACTIVITY DIAGRAM FOR SAVING THE RECORD IN A FILE*

## 3.1 Add New Record

When user clicks on Add New Record button, then a new blank list is shown which will only display the recently added new records for the session (Fig. 8). Here user is able to enter new record for the dictionary. All the text fields where Sindhi script needs to be entered have been designed so that user directly enters the Sindhi text without any settings or modification to the underlying operating system. No Regional and Languages settings are required to use this system to enter Sindhi text. Fig. 9 shows the interface for Sindhi to English Dictionary from where the user enters the record for Sindhi Dictionary.

The system GUI also provides integrated on-screen Sindhi Keyboard (Fig. 10). This on screen integrated Sindhi keyboard provides huge benefit and convenience to the user for entering Sindhi record as there is no further need to install any additional keyboard plug-in for Sindhi language.

## 4. CONCLUSIONS

Due to desperate need of Sindhi computerized dictionaries for different disciplines, this study was initiated to develop a Software architecture through which a Bilingual Sindhi-English dictionary could easily be created. The proposed system has been successfully implemented and evaluated for its proper function and accuracy. The design, implementation and results of this study have been discussed in this article. At present, the system lacks the facility of integrated TTS (Text-To-Speech) system for pronunciation of Sindhi words. The work is under progress in this direction.

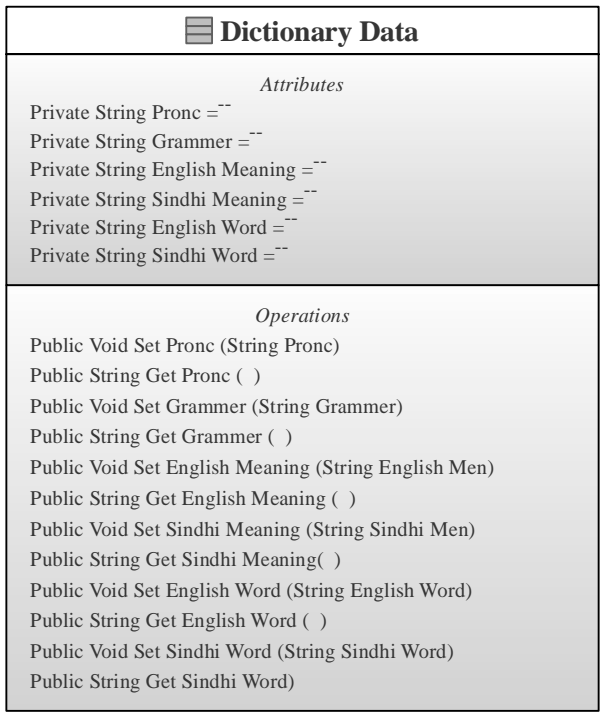

*FIG. 4. CLASS DIAGRAM OF DICTIONARYDATA CLASS*

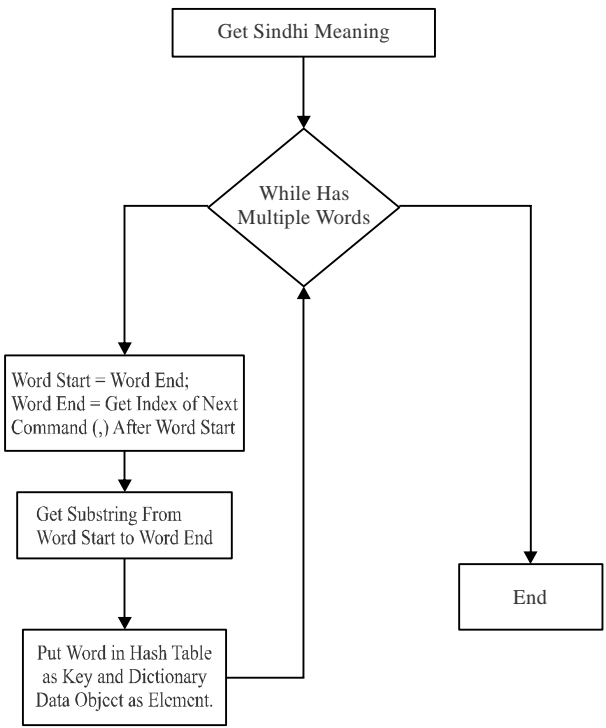

*FIG. 5. FLOW DIAGRAM OF TOKENIZING THE SINDHI MEANING INTO MULTIPLE SINDHI WORDS*

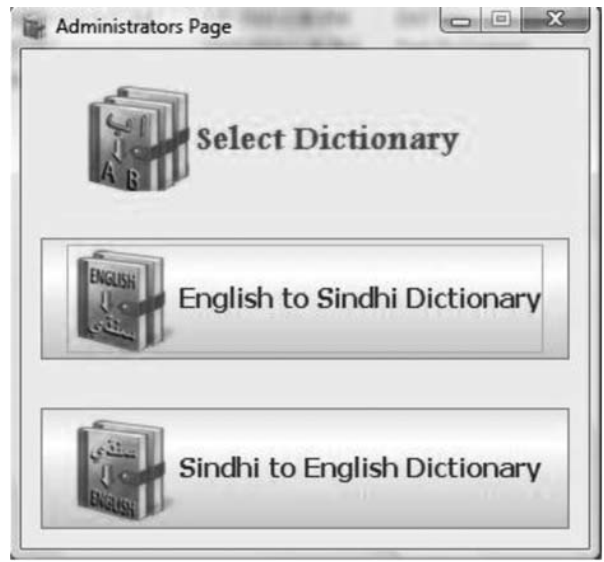

*FIG. 6. SELECTING THE DICTIONARY TYPE: ENGLISH TO SINDHI OR SINDHI TO ENGLISH*

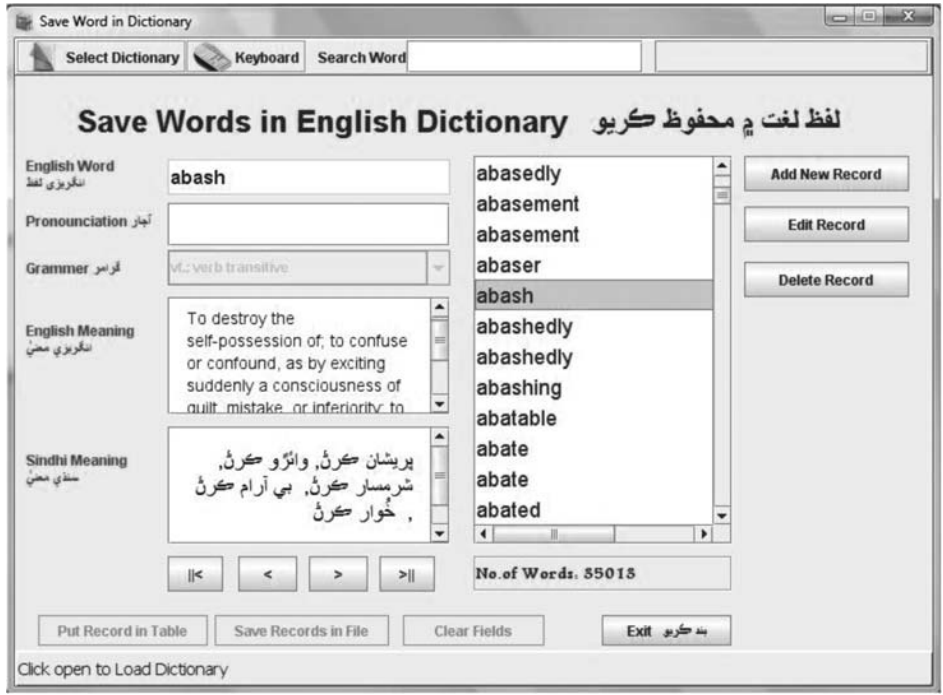

FIG. 7. MAIN GUI OF THE SYSTEM

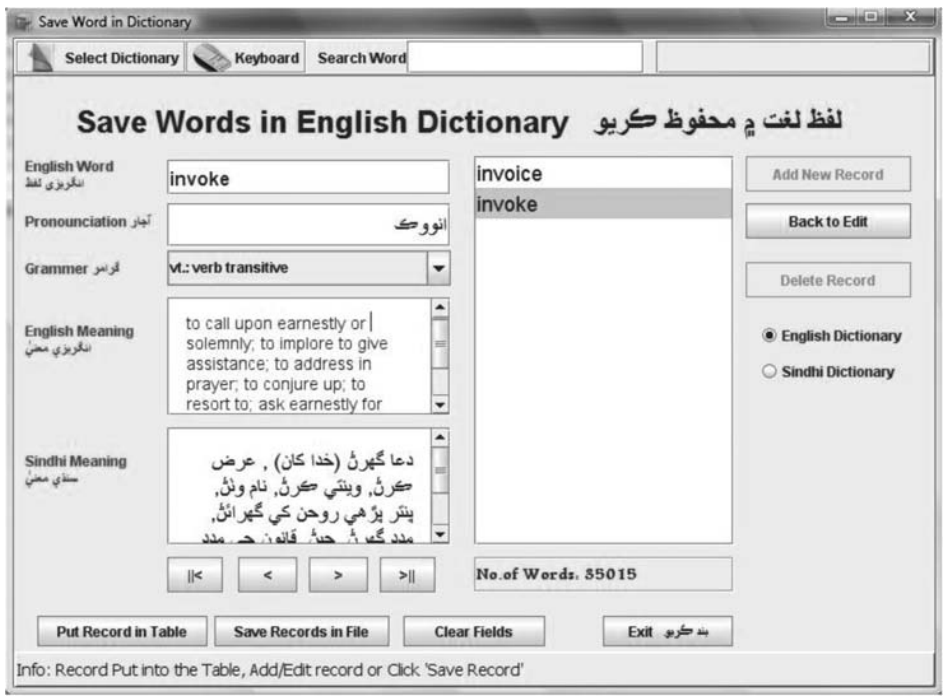

FIG. 8. ADD NEW RECORD GUI FOR ENGLISH TO SINDHI DICTIONARY

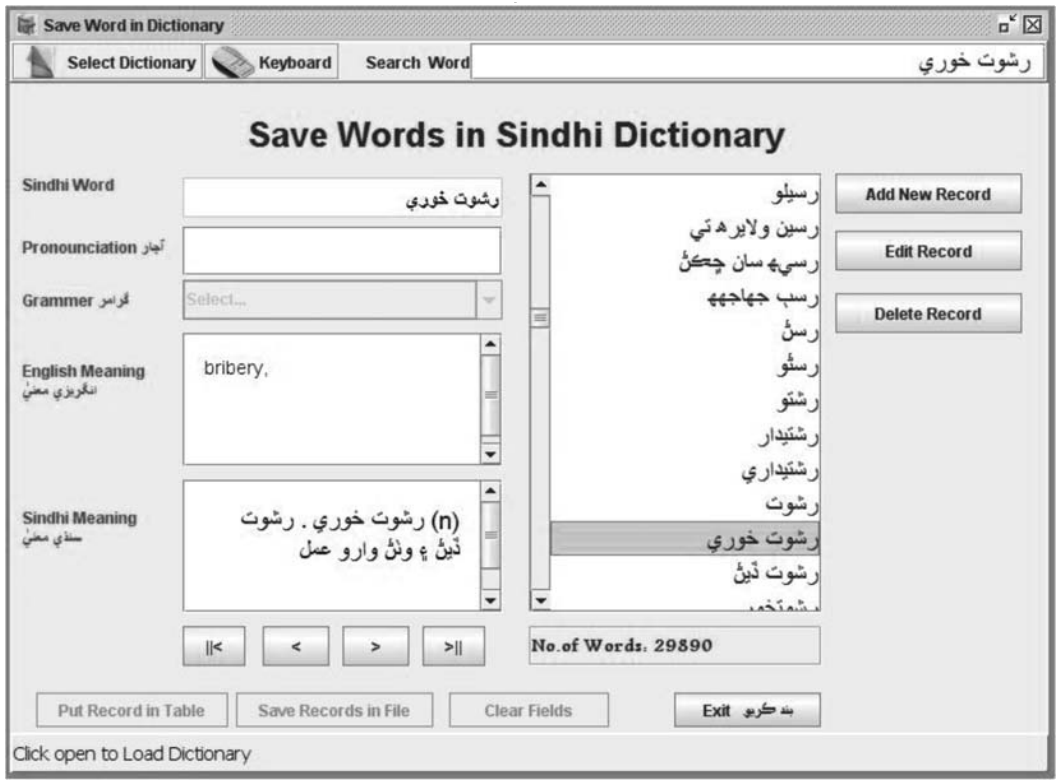

FIG. 9. ADD NEW RECORD GUI FOR SINDHI TO ENGLISH DICTIONARY

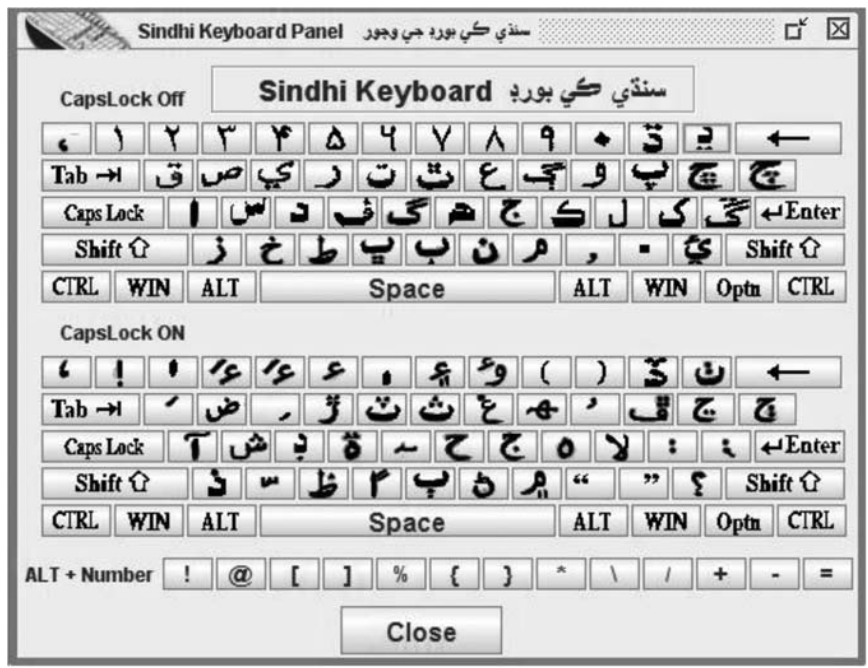

FIG. 10. ON-SCREEN SINDHI KEYBOARD INTEGRATED WITH THE SYSTEM

## ACKNOWLEDGEMENTS

Authors are indebted to acknowledge the valuable input of various Sindhi linguists and also the comments/suggestions of the Internal and External Reviewers, which helped to further improve this work.